\documentclass[sn-basic]{sn-jnl}

\usepackage{graphicx}%
\usepackage{multirow}%
\usepackage{amsmath,amssymb,amsfonts}%
\usepackage{amsthm}%
\usepackage{mathrsfs}%
\usepackage[title]{appendix}%
\usepackage{xcolor}%
\usepackage{textcomp}%
\usepackage{manyfoot}%
\usepackage{booktabs}%
\usepackage{algorithm}%
\usepackage{algorithmicx}%
\usepackage{algpseudocode}%
\usepackage{listings}%

\usepackage{amsmath,amssymb,amsfonts}
\usepackage{algpseudocode,algorithm}
\usepackage{subcaption}
\usepackage[T1]{fontenc}
\usepackage{graphicx}
\usepackage{tabularx}
\usepackage{url}
\usepackage{orcidlink}

\theoremstyle{thmstyleone}%
\theoremstyle{thmstyletwo}%

\theoremstyle{thmstylethree}%

\begin{document}
\title[Improvements of DER and RS]{Improvements of Dark Experience Replay and Reservoir Sampling towards Better Balance between Consolidation and Plasticity}


\author*[1]{\fnm{Taisuke} \sur{Kobayashi} \orcidlink{0000-0002-3760-249X}}\email{kobayashi@nii.ac.jp}

\affil*[1]{\orgname{National Institute of Informatics (NII) / The Graduate University for Advanced Studies (SOKENDAI)}, \orgaddress{\street{2-1-2 Hitotsubashi}, \city{Chiyoda-ku}, \postcode{101-8430}, \state{Tokyo}, \country{Japan}}}

\abstract{%
Continual learning is the one of the most essential abilities for autonomous agents, which can incrementally learn daily-life skills.
For this ultimate goal, a simple but powerful method, dark experience replay (DER), has been proposed recently.
DER mitigates catastrophic forgetting, in which the skills acquired in the past are unintentionally forgotten, by stochastically storing the streaming data in a reservoir sampling (RS) buffer and by relearning them or retaining the past outputs for them.
However, since DER considers multiple objectives, it will not function properly without appropriate weighting of them.
In addition, the ability to retain past outputs inhibits learning if the past outputs are incorrect due to distribution shift or other effects.
This is due to a tradeoff between memory consolidation and plasticity.
The tradeoff is hidden even in the RS buffer, which gradually stops storing new data for new skills in it as data is continuously passed to it.
To alleviate the tradeoff and achieve better balance, this paper proposes improvement strategies to each of DER and RS.
Specifically, DER is improved with automatic adaptation of weights, block of replaying erroneous data, and correction of past outputs.
RS is also improved with generalization of acceptance probability, stratification of plural buffers, and intentional omission of unnecessary data.
These improvements are verified through multiple benchmarks including regression, classification, and reinforcement learning problems.
As a result, the proposed methods achieve steady improvements in learning performance by balancing the memory consolidation and plasticity.
}

\keywords{Continual learning, Consolidation and plasticity, Dark experience replay, Reservoir sampling}

\maketitle
\section{Introduction}

Machine learning technologies have made remarkable progress in recent years~\citep{lecun2015deep,touvron2023llama,kirillov2023segment}, and the basic learning framework is with a huge dataset prepared in advance.
On the other hand, new data is constantly increasing as time goes on, so new skills (or tasks) often arise in addition to ones in the dataset for training.
Alternatively, due to limited computational resources, it is not possible to have all the huge dataset in memory or storage, missing several skills inevitably.
The goal of continual learning (CL) is to incrementally obtain new skills in a machine learning model even in such situations~\citep{parisi2019continual}.
That is, CL needs to train the model using streaming data without preparing a dataset.

In this CL problem setting, catastrophic forgetting (or catastrophic inference), i.e. skills previously obtained are forgotten from the model when new skills are learned, is a major issue~\citep{mcclelland1995there}.
The main objective of research on CL is therefore to alleviate this problem, and three major approaches have been proposed.
\begin{itemize}
    \item Regularization:
    Among the parameters in the model (e.g. weights and biases of neural networks), the essential ones for representing past skills are selected, and then they are regularized for not learning new skills, which can be learned using the remaining ones~\citep{kirkpatrick2017overcoming,farajtabar2020orthogonal}.
    Alternatively, instead of regularization on the parameter space, regularization on the output space of the model can be added to implicitly select and retain the parameters important for the representation of past skills~\citep{titsias2020functional,khan2021knowledge}.
    \item Rehearsal:
    By storing past data in a finite-size buffer, the model is trained with not only streaming data but also data replayed from it, resulting in retaining past skills even when acquiring new skills, as like the standard machine learning with a dataset~\citep{chrysakis2020online,sun2022information}.
    Alternatively, instead of a finite-size buffer, a generative model is be learned additionally, and pseudo-past data can be generated from it and used for learning~\citep{shin2017continual,pomponi2023continual}.
    \item Modularization:
    Each skill is learned mainly in the corresponding module included in the well-designed model, and the modules hardly learn other skills by restricting (especially, freezing) the updates~\citep{kobayashi2020reinforcement,kang2022forget}.
    The amount of representable skills can be increased by adding modules to the model as needed~\citep{li2019learn,ostapenko2021continual}.
\end{itemize}
Note that some CL methods have been proposed that combine these approaches~\citep{buzzega2020dark,daxberger2023improving}.
In addition to this categorization, CL methods can also be classified according to whether or not each data has the label that explicitly represents the corresponding skill (or with or without the timing of changing target skills)~\citep{aljundi2019task,ye2022task}.
Naturally, the methods without label information is more general-purpose and realistic, but it is known that the difficulty of the problem increases significantly.

In the previous CL methods, dark experience replay (DER) (more precisely, DER++ in the original paper) has attracted attention as a simple but very powerful method~\citep{buzzega2020dark}.
DER corresponds to the combination of rehearsal and regularization approaches, whereby past data and corresponding outputs from the model at that time are stored in a buffer, and they are utilized for learning past skills and regularizing the new outputs to maintain past ones, in conjunction with learning on new skills.
Although DER has a very simple implementation that does not require label information, it can mitigate catastrophic forgetting at a high level.
However, behind the simple implementation, DER needs weights for simultaneous optimization of multiple objectives, so it does not function properly without fine-tuning the weights per problem.
Indeed, the original paper showed several benchmark results, but DER had different weights for them.
Furthermore, past outputs do not necessarily represent past skills, so the model tries to maintain outputs even though they may contain errors.
Additional regularization~\citep{zhuo2023continual} and prioritized sampling~\citep{wang2024uncertainty} have been proposed as improvements of DER to reduce the impact of the errors, but they do not eliminate the errors and both have the effect of making the learning process more conservative, thus losing plasticity, which is described below.

In addition, the buffer used in DER is a reservoir sampling (RS) buffer~\citep{vitter1985random}.
While a first-in-first-out (FIFO) buffer, which is widely used in experience replay~\citep{lin1992self,isele2018selective}, always stores new data and discards the oldest data, RS buffer stochastically stores new data and discards data stored.
In this way, the RS buffer can be regarded as sampling a finite number of data with uniform probability from among all data passed so far, remaining data for past skills in the buffer.
However, in other words, the older the data, the more chances there are for it to be included in the buffer, and the smaller the probability and the shorter the duration of recent data being included in the buffer.
Although the value of the buffer can be increased by storing highly informative samples~\citep{sun2022information,aljundi2019gradient,brignac2023improving}, this is not an approach that directly solves this issue.
While some approaches have been proposed to make it easier to accept new data by decaying the storage probability of past data~\citep{cormode2009forward,osborne2014exponential}, excessive decay would undermine the benefits of RS.

Thus, the conventional DER and the RS used in it still have open issues.
In particular, perhaps because DER focuses on resolving catastrophic forgetting, it gives priority to maintain past skills, resulting in the loss of plasticity for efficient acquisition of new skills.
This is related to a tradeoff between stability (consolidation in this paper to distinguish other stability like control/learning stability) and plasticity, which is well-known to be a dilemma faced by humans~\citep{frank2006role,mermillod2013stability}.
Recently, it has been reported that the plasticity of memory should be reconsidered~\citep{dohare2024loss}, so the current trend in which only consolidation is prioritized would be inappropriate.
For example, as the distribution shift problem~\citep{koh2021wilds} suggests, past skills are not always correct, and they must be updated appropriately as the situation changes.
Studies that seek a better balance between consolidation and plasticity often follow an approach of combining two models/structures that are biased in one direction or the other~\citep{jung2023new,kim2023achieving}, which increases computational cost.

Therefore, this study seeks to improve DER and RS in order to obtain a better balance between memory consolidation and plasticity, while not introducing additional models/structures.
As a first contribution, a novel method, so-called A2ER, is proposed with three strategies into DER.
Specifically, the \textit{adaptation} strategy enables auto-tuning of the weights of DER, and appropriately balances learning of new data, learning of past data, and preservation of past outputs.
Next, the \textit{block} strategy suppresses the frequency of replay of past data containing errors to prevent incorrect skills from being consolidated.
Finally, the \textit{correction} strategy corrects errors in past outputs to increase the plasticity.

As a second contribution, a new method, so-called O2S, is proposed with three strategies into RS.
Specifically, the \textit{q-logarithm} strategy generalizes the acceptance probability of the data passed to a RS buffer, and allows specifying the balance between the consolidation and plasticity.
Next, the \textit{plural} strategy prepares multiple RS buffers connected in series and gradually shifts from highly plastic to highly consolidated.
Finally, the \textit{omission} strategy deletes unnecessary past data including errors from the buffers, leaving more important data as long-term memory.

These improvements are numerically verified through multiple benchmarks.
First, to demonstrate the effectiveness of A2ER, an agent with a small buffer trains classification and regression tasks, resulting in that A2ER yields higher accuracy than before, thanks to an appropriate learning balance and improved plasticity.
Similarly, reinforcement learning (RL) tasks can efficiently be accomplished without consolidating incorrectly estimated value functions and past policies.
Next, to demonstrate the effectiveness of O2S, it is confirmed that the underlying \textit{q-logarithm} strategy can specify a balance between the consolidation and plasticity in a classification task in which the distribution shift occurs once.
Finally, O2S achieves higher generalization performance in goal-conditioned RL robotic tasks while saving the amount of data passed to the RS buffers.

\section{Preliminaries}

\subsection{Continual learning}

Let's set up the problem for continual (or lifelong) learning, which is the subject of this study.
First, an agent continuously obtains input data $x_t \in \mathcal{X} \subseteq \mathbb{R}^{|\mathcal{X}|}$ (with $t=1,2,\ldots$ the time step) sequentially from a faced environment.
The corresponding output data $y_t \in \mathcal{Y} \subseteq \mathbb{R}^{|\mathcal{Y}|}$ is predicted.
$y_t$ can be obtained as in supervised learning or estimated by bootstrapping from $x_t$ and other variables as in RL, but the former is assumed here for simplicity of description.
By using a function approximator (e.g. deep neural networks~\citep{lecun2015deep}) with its parameters $\theta$, the following minimization problem is solved.
\begin{align}
    \theta^\ast &= \arg\min_\theta \frac{1}{t} \sum_{\tau=1}^t \mathcal{L}(f(z_\tau), y_\tau)
    \\
    z_\tau &= h_\theta(x_\tau)
    \nonumber
\end{align}
where $h_\theta: \mathcal{X} \mapsto \mathbb{R}^{|\mathcal{Y}|}$ denotes the function approximator to be optimized and $f: \mathbb{R}^{|\mathcal{Y}|} \mapsto \mathcal{Y}$ denotes the fixed mapping function (e.g. sigmoid function).
By setting the appropriate loss function $\mathcal{L}$, $y \simeq f(h_\theta(x))$ is expected.

The difficulty of continual learning stems from the fact that $t$ keeps increasing and its maximum cannot be defined.
In extreme cases, when $t \to \infty$, the above minimization problem cannot be numerically optimized.
In addition, the data size of $\{(x_\tau, y_\tau)\}_{\tau=1}^t$ exceeds the finite computational resources, especially of embodied systems like robots.
Therefore, a FIFO buffer $D^{\mathrm{FIFO}} = \{(x_\tau, y_\tau)\}_{\tau=\min(0, t-N^{\mathrm{FIFO}})+1}^t$ of finite size $N^{\mathrm{FIFO}} \in \mathbb{N}$ is introduced, following the surrogated minimization problem.
\begin{align}
    \theta^\ast = \arg\min_\theta \mathbb{E}_{D^{\mathrm{FIFO}}}[\mathcal{L}(f(h_\theta(x_\tau)), y_\tau)]
    \label{eq:fifo}
\end{align}

This can be minimized by one of stochastic gradient descent methods (e.g.~\citet{ilboudo2023adaterm}).
That is, a batch $B$ of data is randomly extracted from $D^{\mathrm{FIFO}}$, then $\theta$ is updated with its gradient $\nabla_\theta |B|^{-1} \sum_{\tau \in B} \mathcal{L}(f(h_\theta(x_\tau)), y_\tau)$.
Although this solution would allow optimization as in general deep learning, it excludes the past data from optimization when $t > N^{\mathrm{FIFO}}$.
As $t$ increases, therefore, the skills acquired from $\tau \leq \min(0, t-N^{\mathrm{FIFO}})$ are overwritten (unless the data in the FIFO buffer contains equivalent skills).
This overwriting is called catastrophic forgetting~\citep{mcclelland1995there}.

\subsection{Dark experience replay}

Several approaches have been proposed for mitigating catastrophic forgetting, and in this study, dark experience replay (DER)~\citep{buzzega2020dark} (more precisely, DER++ in the original paper) is employed as a baseline with slight modifications.
DER is a simple yet powerful continual learning method that can be regarded as a combination of rehearsal and functional regularization.
In addition to the FIFO buffer, DER introduces an reservoir sampling (RS) buffer~\citep{vitter1985random}, $D^{\mathrm{RS}}$ (with the size $N^{\mathrm{RS}}$) to store past data overflowed from the FIFO buffer.
This RS buffer stochastically stores all data passed so far with equal probability, rather than storing the latest data first as in the FIFO buffer, as described in the next section.
In addition, the feature $z_t$ calculated by $h_\theta$ is additionally stored into the RS buffer together with $(x_t, y_t)$.

Under such a design, the following minimization problem is solved.
\begin{align}
    \theta^\ast = \arg\min_\theta & (1 - \beta) \mathbb{E}_{D^{\mathrm{FIFO}}}[\mathcal{L}(f(h_\theta(x_\tau)), y_\tau)]
    \nonumber \\
    & + \beta \mathbb{E}_{D^{\mathrm{RS}}}[\mathcal{L}(f(h_\theta(x_\tau)), y_\tau)]
    \nonumber \\
    & + \alpha \mathbb{E}_{D^{\mathrm{RS}}}\left[ \frac{1}{2}\|h_\theta(x_\tau) - z_\tau\|_2^2 \right]
    \label{eq:der}
\end{align}
where $\beta \in [0, 1]$ is the coefficient that adjusts the learning priority between the FIFO and RS buffers, and $\alpha \geq 0$ is the strength of the function regularization that tries to preserve past features, which are calculated with old $\theta$.
The second and third terms both randomly select batches from $D^{\mathrm{RS}}$, and while the original implementation was to select each batch independently, in this study, all data are selected simultaneously and divided to two batches so that they do not overlap in order to strengthen the regularization by DER.
Note that this implementation is mostly regarded to be a generalized version of CLEAR~\citep{rolnick2019experience}, which is specialized for RL.

\subsection{Reservoir sampling}

As mentioned briefly before, the RS buffer used in DER has the feature of stochastically storing all data passed so far with equal probability~\citep{vitter1985random}.
For this reason, once the buffer is filled with data, the following algorithm is applied to select the excluded data $d^{\mathrm{del}}$ from the new data passed $n$-th time and $i$-th data in the buffer $\{d_i\}_{i=1}^{N^{\mathrm{RS}}}$.
\begin{align}
    k &\sim \mathcal{U}(1, n)
    \nonumber \\
    d^{\mathrm{del}} &=
    \begin{cases}
        d^\prime & k > N^{\mathrm{RS}}
        \\
        d_k & 1 \leq k \leq N^{\mathrm{RS}}
    \end{cases}
    \label{eq:rs_sample}
\end{align}
where $\mathcal{U}(l, u)$ denotes the integer-type uniform distribution within $[l, u]$, where $l, u \in \mathbb{Z}$ and $l \leq u$.
When excluding data from the buffer, the new data is added to the corresponding index as $d_k = d^\prime$.

In the above algorithm, the probability of accepting the new data $d_n$ passed $n$-th time and the probability that the data still remains in the RS buffer after $n^\prime=1,2,\ldots$ times are as follows:
\begin{align}
    &p(d_n \in D^{\mathrm{RS}}_n)
    \nonumber \\
    &= p(1 \leq k \leq N^{\mathrm{RS}}; k \sim \mathcal{U}(1, n))
    \nonumber \\
    &= \sum_{k=1}^{N^{\mathrm{RS}}} \frac{1}{n} = \frac{N^{\mathrm{RS}}}{n}
    \label{eq:rs_add}
\end{align}
\begin{align}
    &p(d_n \in D^{\mathrm{RS}}_{n+n^\prime})
    \nonumber \\
    &= p(d_n \in D^{\mathrm{RS}}_{n+n^\prime-1}) p(d_n \neq d^{\mathrm{del}}_{n+n^\prime})
    \nonumber \\
    &= p(d_n \in D^{\mathrm{RS}}_{n+n^\prime-1}) \left( \frac{n+n^\prime - N^{\mathrm{RS}}}{n+n^\prime} + \frac{N^{\mathrm{RS}}}{n+n^\prime} \frac{N^{\mathrm{RS}} - 1}{N^{\mathrm{RS}}} \right)
    \nonumber \\
    &= p(d_n \in D^{\mathrm{RS}}_n) \prod_{m=1}^{n^\prime} \frac{n+n^\prime-m}{n+n^\prime+1-m}
    \nonumber \\
    &= \frac{N^{\mathrm{RS}}}{n+n^\prime}
    \label{eq:rs_keep}
\end{align}
These show that the RS buffer stochastically stores all data passed so far with equal probability in inverse proportion to the total number of data passed so far, $n$.
Here, $n$ is sometimes called the reservoir counter~\citep{sun2022information}.

\section{Improvements of DER: A2ER}

\begin{figure*}[tb]
    \centering
    \includegraphics[keepaspectratio=true,width=0.84\linewidth]{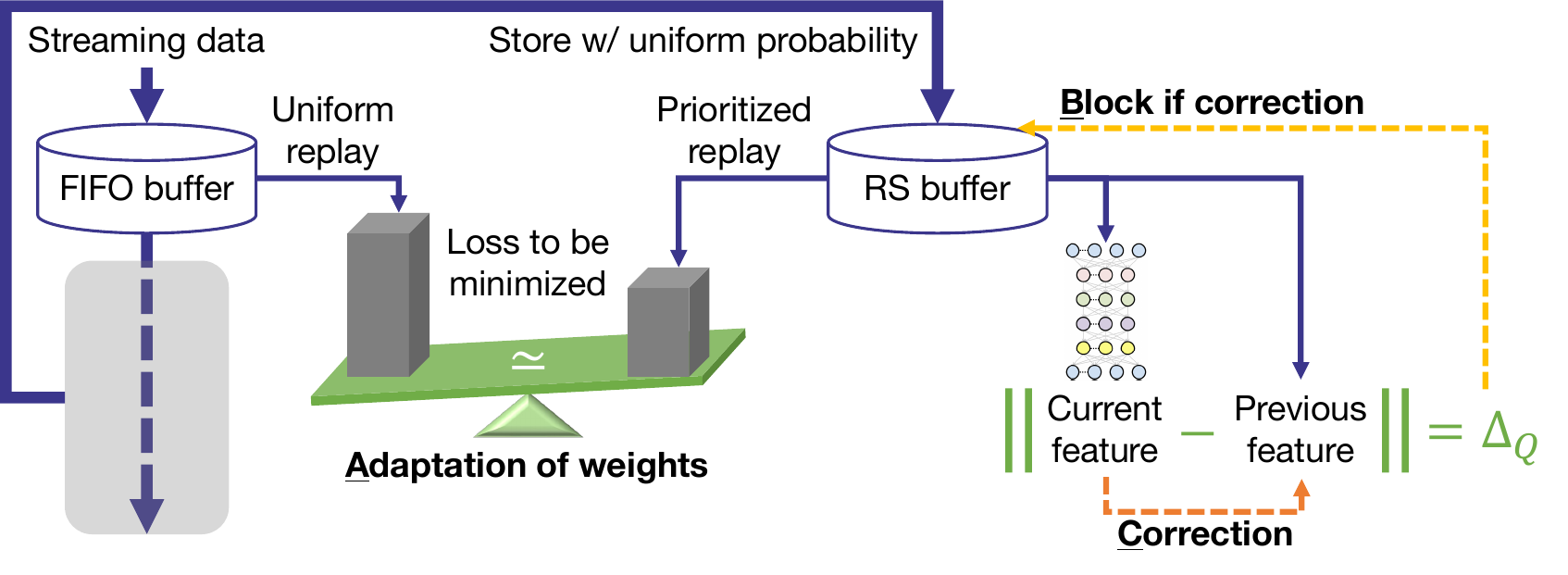}
    \caption{A2ER with three strategies for improving DER}
    \label{fig:concept_a2er}
\end{figure*}

\subsection{Open issues in DER}

DER is an excellent method that can strongly mitigate catastrophic forgetting despite its relatively simple implementation.
However, the hyperparameters $\beta$ and $\alpha$ in DER must be appropriately tuned depending on the target problem to get the best performance.
In fact, the original paper~\citep{buzzega2020dark} tuned different values for each benchmark.
In other words, the balance between the memory consolidation and plasticity must be manually given with a great deal of effort.

In addition, the third term in eq.~\eqref{eq:der}, which corresponds to functional regularization, tries to maintain past features even if they are incorrect.
Alternatively, the features might be correct only for the past situations, not for the situations after distribution shift.
In both cases, this regularization would deteriorate the plasticity in exchange for the consolidation.

To resolve these issues, this paper proposes A2ER, which adds three types of strategies to DER (see Fig.~\ref{fig:concept_a2er}): \textit{adaptation}; \textit{block}; and \textit{correction}.

\subsection{Adaptation of weights}

Here, the \textit{adaptation} strategy attempts to automatially adjust $\beta$ and $\alpha$ in DER, inspired by the recent machine learning formulations~\citep{haarnoja2018soft,stooke2020responsive}.
Specifically, the derivation of DER's minimization problem in eq.~\eqref{eq:der} is reinterpreted as the minimization of eq.~\eqref{eq:fifo} with the following equality constraints, which correspond to the second and third terms in DER.
\begin{align}
    &\mathbb{E}_{D^{\mathrm{RS}}}[\mathcal{L}(f(h_\theta(x_\tau)), y_\tau)] = \mathbb{E}_{D^{\mathrm{FIFO}}}[\mathcal{L}(f(h_\theta(x_\tau)), y_\tau)]
    \\
    &\mathbb{E}_{D^{\mathrm{RS}}}\left[ \frac{1}{2}\|h_\theta(x_\tau) - z_\tau\|_2^2 \right] = \Delta_Q
\end{align}
where $\Delta_Q \geq 0$ is the threshold.

In this paper, this is heuristically designed as variable by the following quantile function $Q$ and updated for each batch calculation.
\begin{align}
    \Delta_Q = (1 - \gamma) \Delta_Q + \gamma Q(\{ \|h_\theta(x_\tau) - z_\tau\|_2^2 \}_{\tau \in B}; \rho)
    \label{eq:threshold}
\end{align}
where $\gamma = |B| / N^{\mathrm{RS}}$ and $\rho \in (0, 1)$ is the position of quantile.

These equality constraints can be returned to the regularization terms via Lagrange multipliers with $\beta$ and $\alpha$.
By considering $\Delta_Q$ as a number independent of $\theta$, this conversion is consistent with eq.~\eqref{eq:der}.
Furthermore, $\beta$ and $\alpha$ can be optimized to satisfy the equality constraints, yielding the follwoing auto-tuning rule.
\begin{align}
    \beta^\ast &= \arg\min_\beta -\beta \{ \mathbb{E}_{D^{\mathrm{RS}}}[\mathcal{L}(f(h_\theta(x_\tau)), y_\tau)]
    \nonumber \\
    &\quad\quad\quad\quad\quad\quad - \mathbb{E}_{D^{\mathrm{FIFO}}}[\mathcal{L}(f(h_\theta(x_\tau)), y_\tau)] \}
    \label{eq:beta}\\
    \alpha^\ast &= \arg\min_\alpha -\alpha \left\{ \mathbb{E}_{D^{\mathrm{RS}}}\left[ \frac{1}{2}\|h_\theta(x_\tau) - z_\tau\|_2^2 \right] - \Delta_Q \right\}
    \label{eq:alpha}
\end{align}
These are also solved by stochastic gradient descent together with the minimization problem in eq.~\eqref{eq:der}.
Note that, although the Lagrange multipliers (i.e. $\beta$ and $\alpha$ in this case) are originally real numbers, their respective domains are restricted to $\beta \in [0, 1]$ and $\alpha \geq 0$.
This restriction can be enforced by sigmoid and softplus functions, respectively, and even with that, $\beta$ and $\alpha$ can be efficiently optimized using the mirror descent method~\citep{beck2003mirror}.

With the above formulation, $\beta$ related to the first constraint brings about a proper balance between the consolidation and plasticity by making the latest data in the FIFO buffer and the past data in the RS buffer have the same level of loss.
The second constraint with $\alpha$ empirically strengthens the consolidation by suppressing excessive changes in features, and increases the plasticity by decreasing the functional regularization after sufficient consolidation.

\subsection{Block of replays and correction of features}

In the second equality constraint, the past data exceeding $\Delta_Q$ are classified as either inappropriate due to distribution shift or with incorrect features due to lack of learning.
Therefore, the \textit{block} strategy restricts the replay probability of that erroneous data like intentional forgetting~\citep{johnson1994processes}, while the \textit{correction} strategy corrects the features in the buffer in case the error is due to lack of learning like memory engram updating~\citep{josselyn2020memory}.
Since these strategies have many processes in common, they are introduced together in this section.

To this end, let's first design how close the error norm $\Delta_\tau = \|h_\theta(x_\tau) - z_\tau\|_2^2 / 2$ of the features for each data should be to $\Delta_Q$ when it exceeds $\Delta_Q$.
The more it deviates from $\Delta_Q$, the more inappropriate the features are considered, so the following $\eta_\tau \in [0, 1]$ is designed to be proportional to the degree of deviation.
\begin{align}
    \eta_\tau &= 1 - \frac{\mathrm{clip}(\Delta_\tau, \Delta_Q, \overline{\Delta}_Q) - \Delta_Q}{\overline{\Delta}_Q - \Delta_Q}
    \\
    \overline{\Delta}_Q &\gets \Delta_Q + \frac{1-\rho}{\rho} \Delta_Q
\end{align}
where $\mathrm{clip}(x, l, u)$ is a function that clips $x$ to fit into $[l, u]$, where $l, u \in \mathbb{R}$ and $l \leq u$.
$\overline{\Delta}_Q$ is an approximate empirical estimate of the maximum $\Delta_\tau$.

\begin{figure}[tb]
    \centering
    \includegraphics[keepaspectratio=true,width=0.96\linewidth]{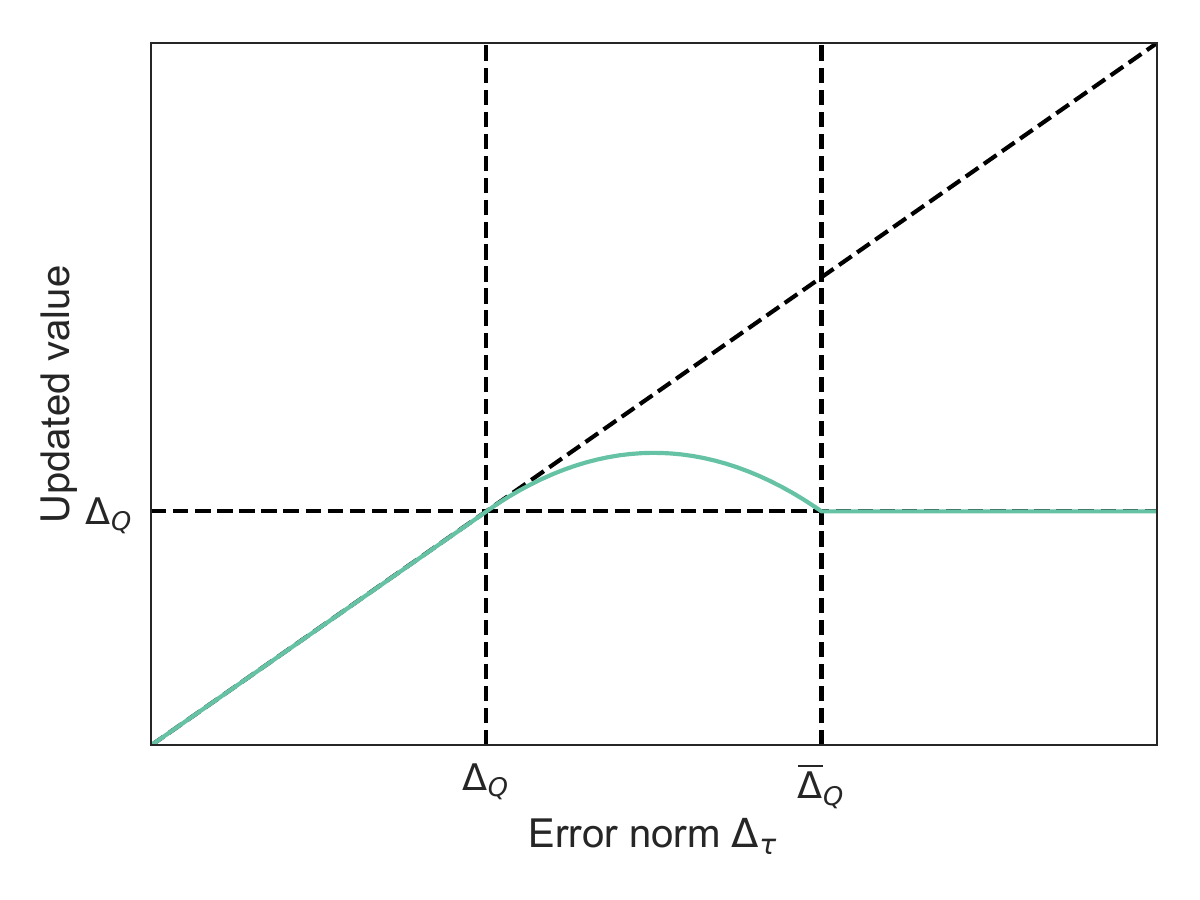}
    \caption{Updated $\Delta_\tau$ using $\eta_\tau$}
    \label{fig:plot_du}
\end{figure}

Using the calculated $\eta_\tau$, let's consider the following update of $\Delta_\tau$ (by correcting $z_\tau$).
\begin{align}
    \Delta_\tau \gets \eta_\tau \Delta_\tau + (1 - \eta_\tau) \Delta_Q
\end{align}
With this desgin, $\Delta_\tau$ at $\Delta_Q \leq \Delta_\tau \leq \overline{\Delta}_Q$ draws a quadratic function with a maximum at $\eta_\tau=1/2$ and $\Delta_Q$ at $\eta_\tau=0,1$, as illustrated in Fig.~\ref{fig:plot_du}.
That is, up to the middle, $\Delta_\tau$ is expected to be minimized by the auto-tuning of $\alpha$, and after that, the behavior is shifted to the correction of $z_\tau$.

Anyway, the correction rate $\gamma_\tau \in [0, 1]$ of $z_\tau$ to $h_\theta(x_\tau)$ is defined as follows:
\begin{align}
    &\eta_\tau \Delta_\tau + (1 - \eta_\tau) \Delta_Q
    \nonumber \\
    &= \frac{1}{2} \|h_\theta(x_\tau) - (1 - \gamma_\tau) z_\tau - \gamma_\tau h_\theta(x_\tau)\|_2^2
    \nonumber \\
    &= \frac{1}{2}\|(1 - \gamma_\tau)\{h_\theta(x_\tau) - z_\tau\}\|_2^2
    \nonumber \\
    &= (1 - \gamma_\tau)^2 \Delta_\tau
    \nonumber \\
    \therefore& \gamma_\tau = 1 - \left\{ \frac{\eta_\tau \Delta_\tau + (1 - \eta_\tau) \Delta_Q}{\Delta_\tau} \right\}^{\frac{1}{2}}
\end{align}

The larger the calculated $\gamma_\tau$ is, the more inappropriate the past features are.
The \textit{block} strategy suppresses the use of the past data itself for replay and learning.
Specifically, although the expectation operation with $D^{\mathrm{RS}}$ in eq.~\eqref{eq:der} replays all data with uniform probability $p_\tau = 1/N^{\mathrm{RS}}$ under normal circumstances, each probability is weighted in the following manner.
\begin{align}
    p_\tau &= \frac{\bar{\gamma}_\tau}{\sum_{\tau^\prime} \bar{\gamma}_{\tau^\prime}}
    \\
    \bar{\gamma}_\tau &\gets (1 - \lambda) \bar{\gamma}_\tau + \lambda (1 - \gamma_\tau)
    \label{eq:gamma}
    \nonumber
\end{align}
where $\lambda \in (0, 1)$ denotes the hyperparameter for the exponential moving average of $1 - \gamma_\tau$.
That is, if the error norm does not become small enough even after learning and the following correction several times, the data will be blocked for replay.

In the \textit{correction} strategy, $z_\tau$ in the buffer is updated using $\gamma_\tau$ as follows:
\begin{align}
    z_\tau \gets (1 - \gamma_\tau) z_\tau + \gamma_\tau h_\theta(x_\tau)
\end{align}
The minimization problem in eq.~\eqref{eq:der} is solved after this correction.
Note that in practice, this can be done by appropriately compensating the loss function associated with $z_\tau$.
That is, $\Delta_\tau - \Delta_Q$ in the update law of $\alpha$ of eq.~\eqref{eq:alpha} is multiplied by $\eta_\tau$ and $\Delta_\tau$ in the update law of $\theta$ of eq.~\eqref{eq:der} is multiplied by $(1 - \gamma_\tau)^2$.
In this way, extra calculations can be omitted by reusing the remaining values in RAM.

\section{Improvements of RS: O2S}

\begin{figure*}[tb]
    \centering
    \includegraphics[keepaspectratio=true,width=0.84\linewidth]{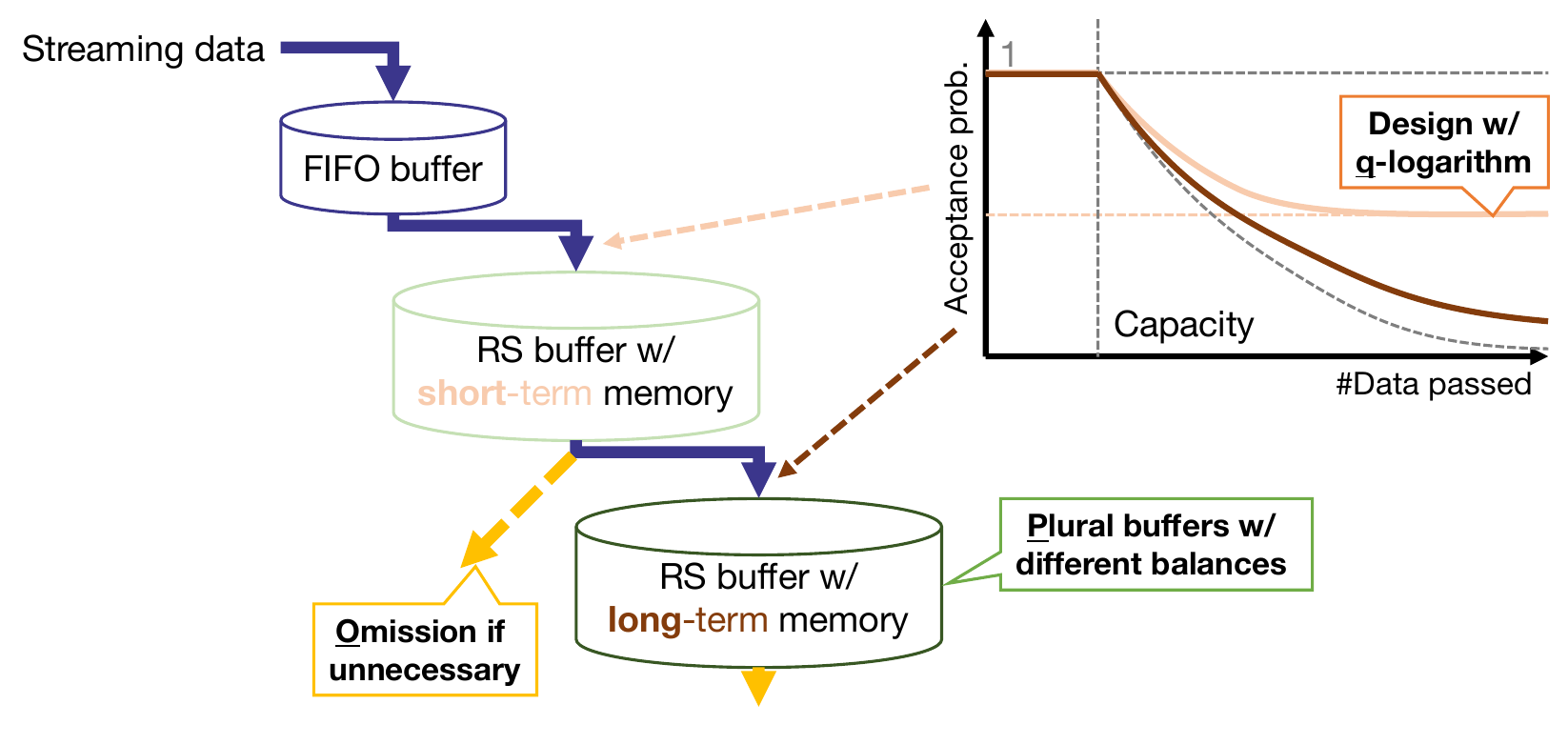}
    \caption{O2S with three strategies for improving RS}
    \label{fig:concept_o2s}
\end{figure*}

\subsection{Open issues in RS}

The RS buffer used in DER accepts and stores data with probability inversely proportional to the reservoir counter $n$, as shown in eqs.~\eqref{eq:rs_add} and~\eqref{eq:rs_keep}.
In other words, when $n$ becomes very large, new data is almost never accepted into the RS buffer, although if accepted once, it can be stored for a long time.
This characteristic is effective in preventing past data from being discarded and improving the consolidation, but it impairs the plasticity to adapt to distributional shift.

In addition, data have varying degrees of importance as suggested by the \textit{block} strategy, and it is undesirable to pass unnecessary data to the RS buffer, as like the literature~\citep{sun2022information}.
In other words, if unnecessary data is stored in the RS buffer, it can impede learning.
Furthermore, just passing it to the RS buffer increases $n$, which decreases the acceptance rate of necessary data passed later.

To resolve these issues, this paper proposes O2S, which adds three types of strategies to RS (see Fig.~\ref{fig:concept_o2s}): \textit{q-logarithm}; \textit{plural}; and \textit{omission}.

\subsection{Generalization of acceptance rate using q-logarithm}

First of all, the acceptance rate is generalized in order to balance between the consolidation and plasticity.
Specifically, a monotonically nondecreasing transformation $f: \mathbb{Z} \mapsto \mathbb{Z}$ is designed for generating random numbers at $n > N^{\mathrm{RS}}$, as $k \sim \mathcal{U}(1, f(n))$ in eq.~\eqref{eq:rs_sample}.
Note that the reason for making $f$ monotonically nondecreasing is that $f(n)$ is also interpreted as a kind of counter.
The acceptance rate of the $n$-th new data $d_n$ is given as follows:
\begin{align}
    p(d_n \in D^{\mathrm{RS}}_n) = \frac{N^{\mathrm{RS}}}{f(n)}
    \label{eq:rs_add_gen}
\end{align}
In order for this to satisfy the definition of probability and still have continuity with $n \leq N^{\mathrm{RS}}$ when the acceptance rate is $1$, $\lim_{n \to N^{\mathrm{RS}}} f(n) = N^{\mathrm{RS}}$ holds.

Considering the probability that $d_n$ remains in the RS buffer for the $n+n^\prime$-th time, one term cannot be canceled out, although the derivation procedure is similar to eq.~\eqref{eq:rs_keep}.
\begin{align}
    &p(d_n \in D^{\mathrm{RS}}_{n+n^\prime})
    \nonumber \\
    &= p(d_n \in D^{\mathrm{RS}}_{n+n^\prime-1}) p(d_n \neq d^{\mathrm{del}}_{n+n^\prime})
    \nonumber \\
    &= p(d_n \in D^{\mathrm{RS}}_{n+n^\prime-1})
    \nonumber \\
    &\quad \times \left( \frac{f(n+n^\prime) - N^{\mathrm{RS}}}{f(n+n^\prime)} + \frac{N^{\mathrm{RS}}}{f(n+n^\prime)} \frac{N^{\mathrm{RS}} - 1}{N^{\mathrm{RS}}} \right)
    \nonumber \\
    &= p(d_n \in D^{\mathrm{RS}}_{n+n^\prime-1}) \left( \frac{f(n+n^\prime) - 1}{f(n+n^\prime)} \right)
    \nonumber \\
    &= \frac{N^{\mathrm{RS}}}{f(n+n^\prime)} \prod_{m=1}^{n^\prime} \frac{f(n+m)-1}{f(n+m-1)}
    \label{eq:rs_keep_gen}
\end{align}
Here, the first term corresponds to the conventional probability of $f(n)=n$, but the second term, i.e. the total product operation, modifies it.

Even with such a modification, the definition probability must be satisfied.
To this end, the condition for designing $f$ is revealed by focusing on inside of the second term.
\begin{align}
    \frac{f(n+m)-1}{f(n+m-1)} &= \frac{\Delta f(n+m) + f(n+m-1) - 1}{f(n+m-1)}
    \nonumber \\
    &= 1 + \frac{\Delta f(n+m) - 1}{f(n+m-1)}
\end{align}
where $\Delta f(n) = f(n) - f(n-1)$.
If this term is sometimes larger than $1$, the probability may exceed $1$ and deviate from the definition.
Therefore, the sufficient condition $\Delta f(n) \leq 1$ should be satisfied.

From the above, the generalized counter $f(n)$ is a function satisfying the following conditions.
\begin{align}
    \begin{split}
        \lim_{n \to N^{\mathrm{RS}}} f(n) = N^{\mathrm{RS}}
        \\
        0 \leq \Delta f(n) \leq 1
    \end{split}
\end{align}
When $\Delta f(n) = 1$, new data and past data are equally likely to be included in the buffer after updating, which makes it easier to store past data and provides the consolidation.
When $\Delta f(n) = 0$, the probability of past data remaining in the buffer decays exponentially due to the total product operation according to $f(n) = c$ at convergence, thus facilitating the acceptance of new data and providing the plasticity.
In other words, if $\Delta f(n)$ can be specified and varied to a good degree, it should be possible to achieve a balance between the consolidation and plasticity.

There are several possible candidates for functions that satisfy these conditions, but this paper introduces one that makes use of the following $q$-logarithmic function~\citep{tsallis1988possible} as the \textit{q-logarithm} strategy.
\begin{align}
    f_q(n) &= \min(n, N^{\mathrm{RS}})
    \nonumber \\
    &\quad + \left\lfloor N^{\mathrm{RS}} \ln_q \left(1 + \frac{\max(0, n - N^{\mathrm{RS}})}{N^{\mathrm{RS}}} \right) \right\rfloor
    \label{eq:qlog}\\
    \ln_q(x) &=
    \begin{cases}
        \ln(x) & q = 1
        \\
        \frac{x^{1-q}-1}{1-q} & q \neq 1
    \end{cases}
    \nonumber
\end{align}
where $q \in [0, 2]$ denotes the hyperparameter to balance the consolidation and plasticity.

\begin{figure}[tb]
    \centering
    \includegraphics[keepaspectratio=true,width=0.96\linewidth]{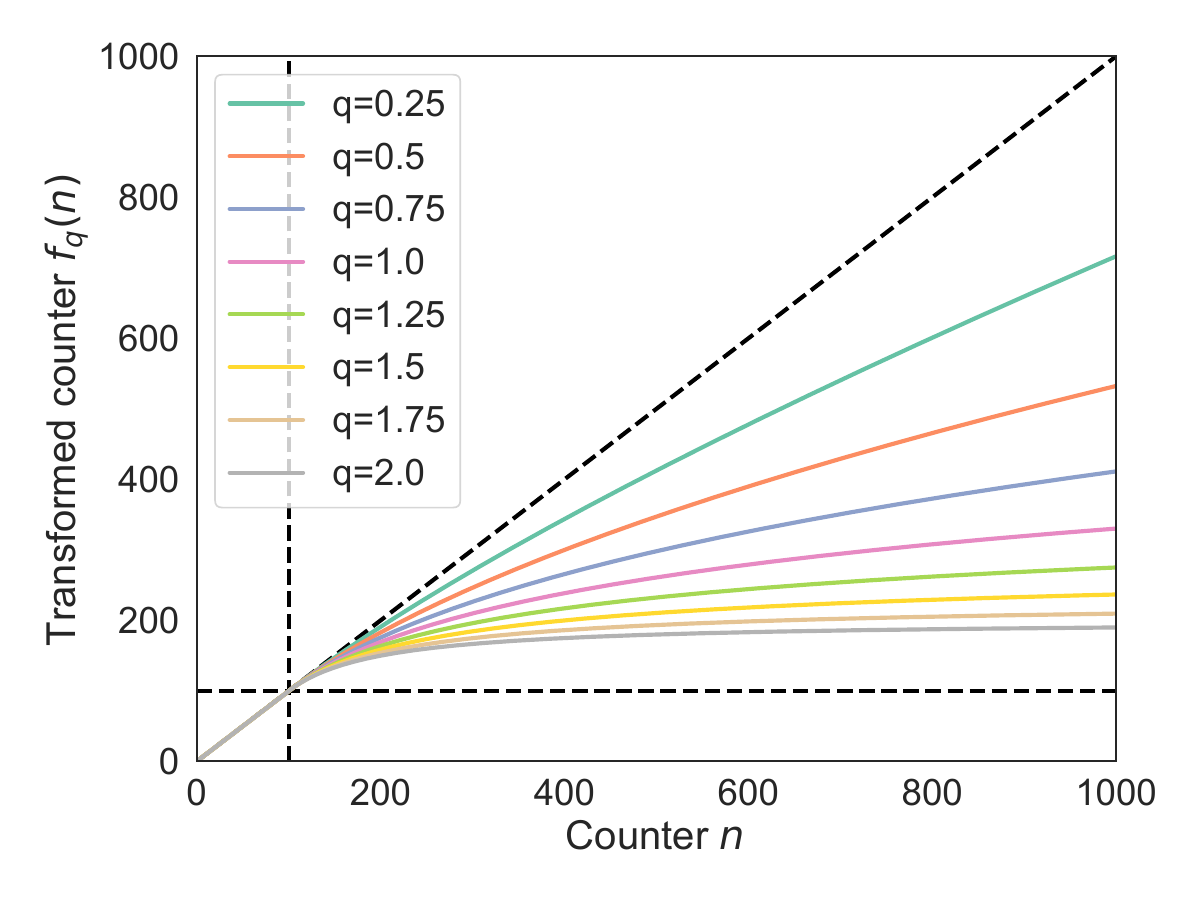}
    \caption{Example of $f_q(n)$ with $N^{\mathrm{RS}}=100$}
    \label{fig:plot_fq}
\end{figure}

$f_q(n)$ is visualized in Fig.~\ref{fig:plot_fq} for a better understanding.
As can be seen in the figure, for $q_1 > q_2$, $f_{q_1}(n) \leq f_{q_2}(n)$ holds.
For $q = 0$, $f_q(n)$ reverts to the conventional RS with $f_{q=0}(n) = n$, which has the highest consolidation.
For $0 \leq q \leq 1$, $\lim_{n \to \infty}f_q(n) = \infty$, it will eventually stop accepting new data and store past data, although there is a time lag until convergence.
On the other hand, for $q > 1$, a bounded upper bound exists and converges to $\lim_{n \to \infty}f_q(n) = N^{\mathrm{RS}} + N^{\mathrm{RS}} / (q - 1)$, so new data is accepted with constant probability and past data decays exponentially, increasing the plasticity.
Note that although $q \to \infty$ is acceptable, the acceptance rate of new data converges to $0.5$ with $q = 2$, which would be of the sufficient plasticity, this paper restricts $q$ within $[0, 2]$, as defined before.

Thus, the proposed design is such that small $q$ increases the consolidation and large $q$ increases the plasticity, being able to continuously adjust the balance between them.
Note that other possible simple designs are examined in \ref{app:design}.

\subsection{Plural buffers and omission of data passing}

Although the designed counter $f_q$ can indeed adjust the balance between the consolidation and plasticity, the tradeoff itself has not been eliminated and may lead to a halfway performance unless it is adjusted appropriately for the target domain.
To alleviate this limitation, the \textit{plural} strategy introduces a layered structure with multiple RS buffers with different configurations, as like human's short- and long-term memory structure~\citep{cowan2008differences}.
Furthermore, when passing data between multiple RS buffers, the importance of the data is calculated in terms of priority of sampling.
The \textit{omission} strategy, therefore, utilizes that information to determine whether the data should be passed to the next buffer.

\begin{figure}[tb]
    \centering
    \includegraphics[keepaspectratio=true,width=0.96\linewidth]{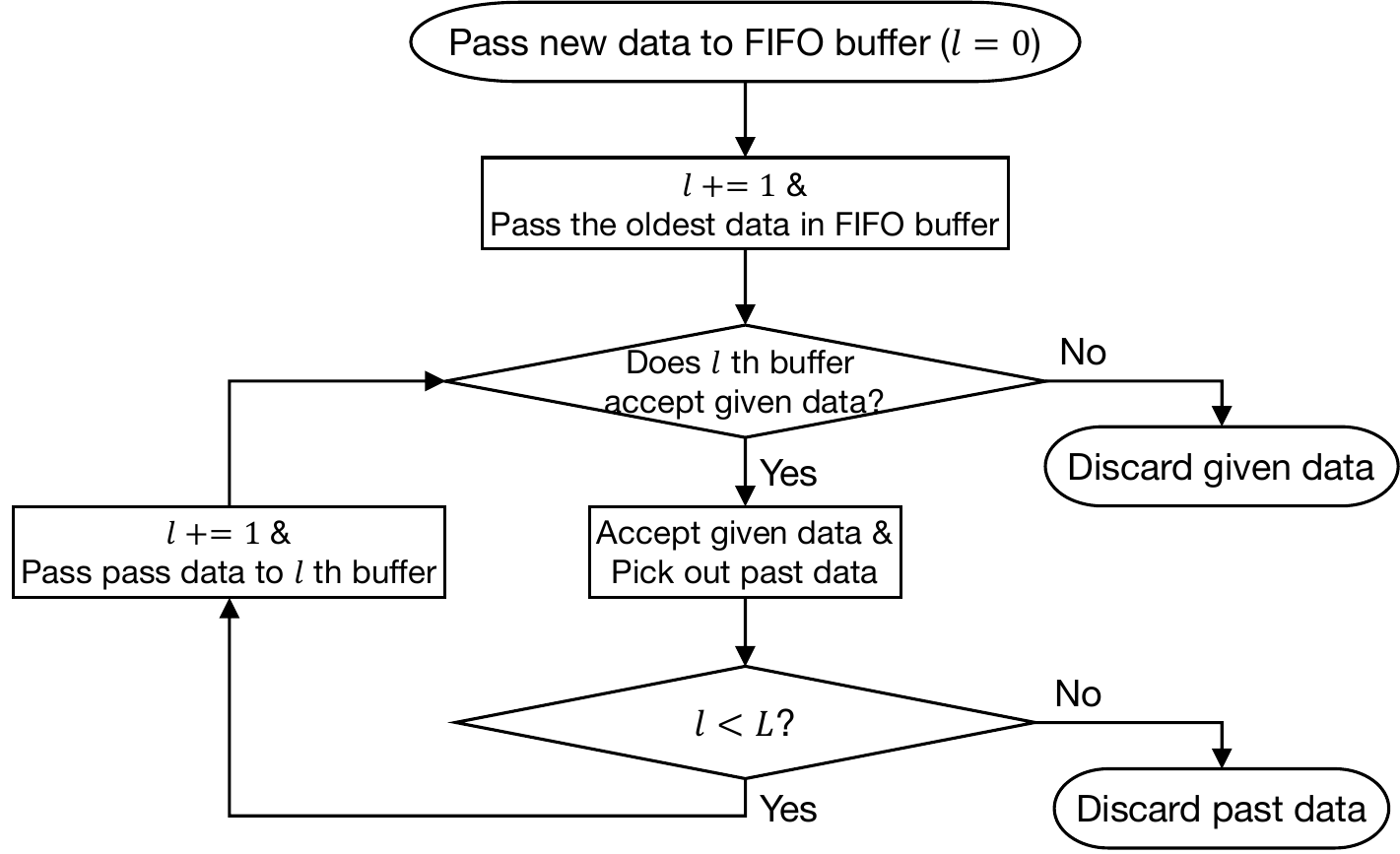}
    \caption{Data processing in \textit{plural} strategy with $L$ RS buffers}
    \label{fig:flow_plural}
\end{figure}

First, the \textit{plural} strategy prepares $L \in \mathbb{N}$ layers of serially connected RS buffers $\{D^{\mathrm{RS}}_l\}_{l=1}^L$ with the respective sizes $N^{\mathrm{RS}}_l$.
They have their own counters $n_l$ and the balance $q_l$ for $f_q$ in eq.~\eqref{eq:qlog}.
Here, the FIFO buffer is regarded as the special buffer at $l=0$.
The data processing is shown in Fig.~\ref{fig:flow_plural}: when the past data in the $l-1$-th buffer is discarded, it is passed to the $l$-th buffer; when the new data passed to $l-1$-th buffer is not accepted, it is not passed to the $l$-th buffer.
If $q_l$ is small and the consolidation is prioritized in the shallow layer of $l \simeq 1$, it rarely discards the past data, preventing the flow to subsequent layers.
Therefore, for $l_1 < l_2$, $q_{l_1} \geq q_{l_2}$ is desired.
In this way, buffers in the deeper layers of $l \simeq L$ will have a longer time scale, since new data will be passed less frequently.
Note that the batch $B$ for training with eq.~\eqref{eq:der} is given as the sum of sub batches (with $B/L$ the size for each) sampled from all the RS buffers.

Second, the \textit{omission} strategy intercepts the data passing from the $l-1$-th to $l$-th buffers.
As mentioned above, each data has the importance represented by the priority of replay $\bar{\gamma}$.
In other words, as data that has already been or will be blocked from being replayed tends to waste the buffer space, the buffer should be fully utilized without advancing its counter by discarding the data before passing data to it.
While converting $\bar{\gamma}$ to the rejection probabilities in the respective buffers $p_{l-1,l}^\mathrm{rej}$, data at time $\tau$ is passed to the $l$-th buffer as $d^\prime_l$ with the following rejection probability $p^\mathrm{rej}$.
\begin{align}
    p^\mathrm{rej} &= 1 - (1 - p_{l-1}^\mathrm{rej}) (1 - p_{l}^\mathrm{rej})
    \\
    p_{l-1,l}^\mathrm{rej} &= \left( \frac{\bar{\gamma}_{l-1,l}^\mathrm{max} - \bar{\gamma}_\tau}{\bar{\gamma}_{l-1,l}^\mathrm{max} - \bar{\gamma}_{l-1,l}^\mathrm{min}} \right)^\nu
    \nonumber
\end{align}
where $\bar{\gamma}_{l-1,l}^\mathrm{max}$ and $\bar{\gamma}_{l-1,l}^\mathrm{min}$ denote the maximum and minimum values in the $l-1$- or $l$-th buffer and $\bar{\gamma}_\tau$, respectively.
$\nu \geq 0$ adjusts the rejection rate, but it is difficult to be tuned intuitively.
Instead, this paper specifies the rejection probability $\zeta \in [0, 1]$ at $p_{l-1}^\mathrm{rej} = p_{l}^\mathrm{rej} = 0.5^\nu$ (with the intermediate priority), resulting in $\nu = - \ln(1 - \sqrt{1-\zeta})/\ln 2$.
Note that as the FIFO buffer has no $\bar{\gamma}$, the data passing from $l=0$ to $l=1$ is always accepted with $p_0^\mathrm{rej} = 0$.
In addition, for numerical stability, $p_{\cdot}^\mathrm{rej} = 0$ if $\bar{\gamma}_{\cdot}^\mathrm{max} - \bar{\gamma}_{\cdot}^\mathrm{min}$ is smaller than $\epsilon$ (in this paper, $10^{-5}$).

\section{Numerical verifications}

\subsection{Common setup}

To validate the effectiveness of the proposed A2ER and O2S, multiple numerical benchmarks are performed.
The basic hyperparameter settings are the same for all benchmarks.
That is, as hyperparameters given by DER~\citep{buzzega2020dark}, the size of the FIFO buffer is 512, the size of the RS buffer is also 512 (in total, even if the \textit{plural} strategy is applied), and the batch size of data replayed from each buffer is 32.
The initial value of $\alpha$ in eq.~\eqref{eq:der} is $1$, and the initial value of $\beta$ is $0.5$.
Note that the training frequency relative to appending data and the number of batches replayed per training are specified for each benchmark.
In addition, AdaTerm~\citep{ilboudo2023adaterm} is employed by default for optimization with stochastic gradient descent.

As hyperparameters involved in the proposed method, $\rho$ in eq.~\eqref{eq:threshold} of the quantile function seems to be of particular importance, and its influence was investigated in \ref{app:quantile}.
As a result, $\rho=0.5$ (i.e. the median value) is appropriate.
Although there is room for adjustment of $\lambda$, which is defined in eq.~\eqref{eq:gamma}, it is set to $0.5$, the center of its domain, for simplicity.
$q$ required for eq.~\eqref{eq:qlog} is set to $1$ for the last buffer, based on the comparisons with other designs in \ref{app:design}.
For simplicity, $L=2$ is used for the \textit{plural} strategy in this paper, and for the first buffer, $q=1.5$ is used to emphasize plasticity.
Finally, the probability of rejecting unnecessary data when transferring data between buffers is empirically set to $\zeta=0.2$ so that excessive counter growth can be suppressed to some extent by rejection.

The above settings are summarized in Table~\ref{tab:param}.
Although there is room for fine-tuning these settings depending on the problem, it seems not to be very important, since the effectiveness of the proposed method has been confirmed for many benchmark problems, as shown below.

\begin{table}[tb]
    \caption{Parameter configuration}
    \label{tab:param}
    \centering
    \begin{tabular}{ccc}
        \hline\hline
        Symbol & Meaning & Value
        \\
        \hline
        $N^{\mathrm{FIFO}}$ & The size of FIFO buffer & $512$
        \\
        $N^{\mathrm{RS}}$ & The size of RS buffer(s) & $512$
        \\
        $B$ & Batch size & $32$
        \\
        $\alpha^{\mathrm{ini}}$ & Initial $\alpha$ in DER & $1$
        \\
        $\beta^{\mathrm{ini}}$ & Initial $\beta$ in DER & $0.5$
        \\
        $\rho$ & Quantile for computing threshold & $0.5$
        \\
        $\lambda$ & Smoothness of data priority & $0.5$
        \\
        $q$ & Balance of generalized counter(s) & $(1.5, 1)$
        \\
        $\zeta$ & Rejection probability & $0.2$
        \\
        \hline\hline
    \end{tabular}
\end{table}

\subsection{Results of A2ER}
\label{subsec:res_a2er}

\subsubsection{Toy problems}

\begin{figure}[tb]
    \begin{subfigure}[b]{0.48\linewidth}
        \centering
        \includegraphics[keepaspectratio=true,width=\linewidth]{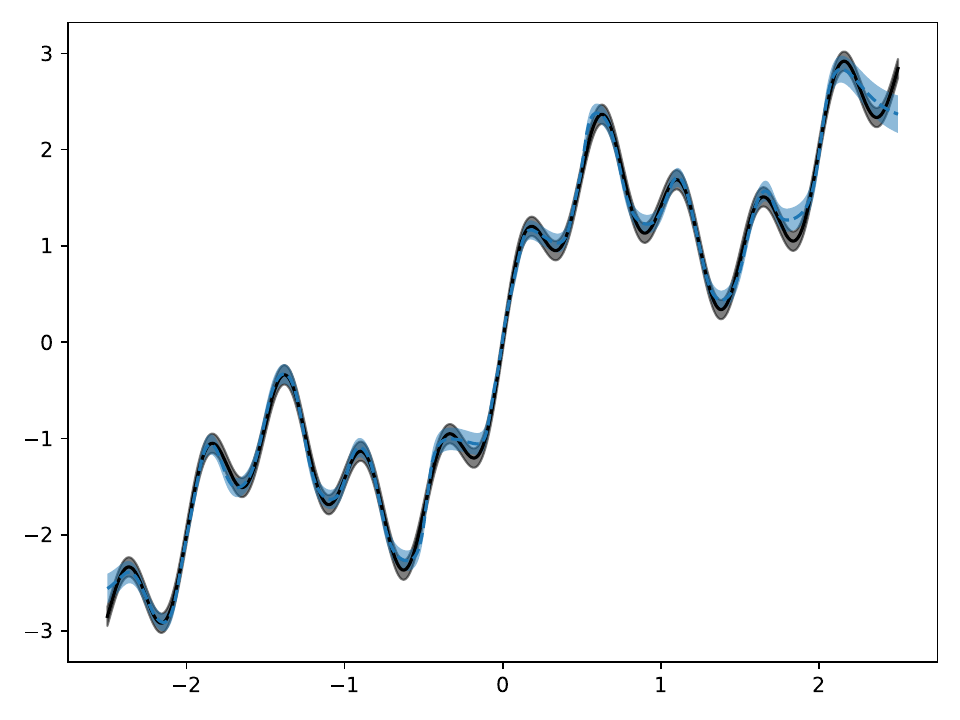}
        \subcaption{Regression}
        \label{fig:prob_regression}
    \end{subfigure}
    \begin{subfigure}[b]{0.48\linewidth}
        \centering
        \includegraphics[keepaspectratio=true,width=\linewidth]{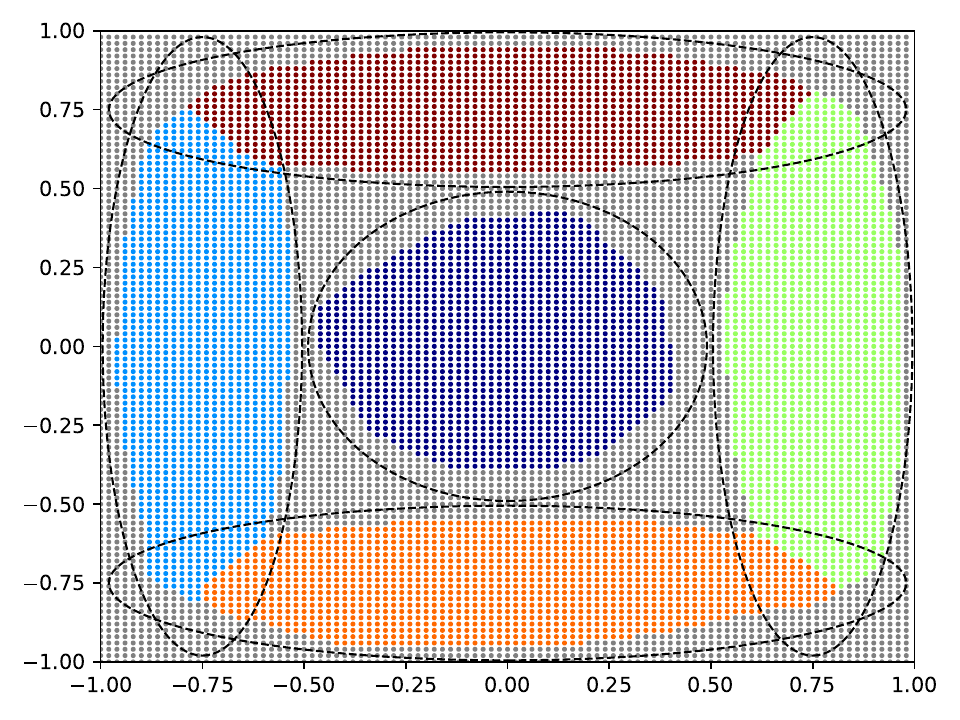}
        \subcaption{Classification}
        \label{fig:prob_classification}
    \end{subfigure}
    \caption{Examples of toy problems}
    \label{fig:prob_toy}
\end{figure}

\begin{table*}[tb]
    \caption{Results of toy problems for evaluating A2ER:
    the average of 20 trials in each condition was weighted by ranks, giving priority to the worst case;
    top-2's methods were noted in bold.
    }
    \label{tab:result_a2er_toy}
    \centering
    \tiny
    \begin{tabular}{l|cccc|cccc}
        \hline\hline
        Method & \multicolumn{4}{c|}{Regression (KLD: smaller is better)} & \multicolumn{4}{c}{Classification (ACC: larger is better)}
        \\
         & R1 & R2 & R3 & R4 & C1 & C2 & C3 & C4
        \\
        \hline
        DER & 0.84 & 4.00 & 7.24 & 2.23 & 84.12 & 73.25 & 80.97 & 84.58
        \\
         & [0.64, 0.92] & [3.71, 4.18] & [6.81, 7.48] & [1.98, 2.32] & [82.69, 85.82] & [71.37, 75.64] & [77.81, 83.91] & [82.70, 86.82]
        \\
        -Aa & 0.49 & \textbf{2.83} & 7.72 & 1.72 & 89.99 & 85.82 & 89.88 & 92.56
        \\
         & [0.28, 0.73] & [2.29, 3.36] & [4.91, 13.36] & [1.51, 1.86] & [88.83, 91.57] & [85.18, 87.45] & [87.85, 92.26] & [91.67, 93.98]
        \\
        -Ab & \textbf{0.40} & 3.07 & \textbf{6.59} & \textbf{1.63} & 90.00 & 85.91 & 90.74 & \textbf{96.45}
        \\
         & [0.26, 0.49] & [2.50, 4.35] & [4.43, 8.29] & [1.44, 1.93] & [88.30, 92.21] & [84.92, 87.90] & [89.61, 93.17] & [95.94, 97.19]
        \\
        -B & \textbf{0.36} & 2.90 & 7.12 & \textbf{1.69} & \textbf{91.33} & \textbf{87.63} & \textbf{91.77} & 95.79
        \\
         & [0.28, 0.45] & [2.36, 3.13] & [4.15, 12.07] & [1.44, 1.85] & [90.61, 92.76] & [86.71, 88.34] & [90.94, 93.40] & [95.29, 96.63]
        \\
        -C & 1.56 & 6.19 & 10.44 & 2.14 & 65.99 & 39.68 & 62.24 & 64.78
        \\
         & [0.83, 1.96] & [5.56, 6.41] & [9.50, 10.66] & [1.81, 2.34] & [63.38, 71.11] & [35.69, 47.84] & [60.25, 67.82] & [59.90, 71.34]
        \\
        A2ER & 0.42 & \textbf{2.87} & \textbf{7.03} & 1.71 & \textbf{91.26} & \textbf{87.52} & \textbf{91.28} & \textbf{95.82}
        \\
         & [0.30, 0.49] & [2.37, 3.59] & [4.44, 10.68] & [1.39, 1.78] & [90.07, 92.70] & [87.03, 88.70] & [89.89, 93.39] & [95.33, 96.85]
        \\
        \hline\hline
    \end{tabular}
\end{table*}

First, to validate the effectiveness of A2ER, simple regression and classification problems are conducted, as shown in Fig.~\ref{fig:prob_toy}.
For both toy problems, a neural network model consisting of two full-connected layers with 32 neurons is trained.
The model outputs a normal distribution for the regression problem and a categorical distribution for the classification problem.
In both cases, the model is optimized by minimizing the negative log-likelihood with respect to the supervised data as the loss function.

The regression problem aims to predict the one-dimensional output of a composite function of up to seven sine waves with different phase, frequency, and amplitude for a one-dimensional input $[-2.5, 2.5]$.
During training, the outputs with white noise (SD is $0.1$) are obtained as the input increases in $0.001$ increments from the minimum value, and the input-output pairs are passed to the model in sequence.
One cycle has 5,000 data, exceeding the buffer size, and it is repeated five times in total to achieve sufficient accuracy.
The training timing is after the addition of 16 data points, and up to 16 batches are replayed from each buffer in a single training session.
This problem is evaluated by the sum of Kullback-Leibler divergences (KLD) between the predicted and true distributions.

The classification problem aims to predict which component of a Gaussian mixture distribution with up to 16 components placed on a two-dimensional input $[-1, 1]^2$ space belongs or does not belong to any of them.
During training, the input space is divided into grids every $0.02$ with white noise (SD is $0.01$), and the outputs are the indices with the maximum likelihood among all components.
The threshold is set to $0.3^2$, and if all likelihoods are less than this threshold, the indices are assigned to different ones, meaning outliers.
This cycle is repeated five times for a total of 10,000 data points to achieve correct classification.
The training timing is after the addition of 32 data points, and up to 16 batches are replayed from each buffer in a single training session.
This problem is evaluated by the classification accuracy (ACC).

Under the above setup, four types of functions (i.e. sine waves and Gaussian mixture distributions) are prepared.
They are trained with the following conditions, each of which is evaluated statistically with 20 different random seeds.
\begin{itemize}
    \item \textit{DER}: the conventional method
    \item \textit{-Aa}: without the \textit{adaptation} strategy for $\alpha$
    \item \textit{-Ab}: without the \textit{adaptation} strategy for $\beta$
    \item \textit{-B}: without the \textit{block} strategy
    \item \textit{-C}: without the \textit{correction} strategy
    \item \textit{A2ER}: the proposed method
\end{itemize}
Since the purpose of this section is to evaluate A2ER, the conventional implementation of RS, not O2S, is employed.

The learning results are summarized in Table~\ref{tab:result_a2er_toy}.
Here, in order to evaluate robustness to problems, a weighted average based on the ranks of the evaluation values is used to prioritize the worst cases.
The conventional method, DER, was never in the top-2, perhaps because it was not fine-tuned and did not perform well enough.
The performance without the \textit{correction} strategy was significantly worse than that of the conventional method.
This may be due to the fact that the \textit{adaptation} strategy gives too much weight $\alpha$ without the reduction of the threshold $\Delta_Q$ by the \textit{correction} strategy, inhibiting learning.

In the other conditions where the \textit{correction} strategy was added, the performances were better than that of the conventional method.
In particular, the proposed method, A2ER, achieved the highest number of top-2 entries.
However, the performance without the \textit{block} strategy was similar, and its usefulness was not confirmed from these benchmarks.
On the other hand, by focusing on the case without the \textit{adaptation} strategy, we can find that the automatic adjustment of $\alpha$ is particularly effective.
When $\beta$ was auto-tuned, the performance was degraded mainly in the classifiation problems.
In fact, while $\beta$ in A2ER stayed around $0.5$ for the regression problems, it temporarily increased to over $0.8$ for the classification problems.
That means, the automatic adjustment of $\beta$ improved performance by correcting improperly initialized $\beta$ to an appropriate value.

The above results indicate that at least the \textit{adaptation} and \textit{correction} strategies of the proposed method are necessary to improve the performance of DER.
As for the \textit{block} strategy, its performance might be activated only when incorrect data remain in the buffers due to, for example, shifts in data generative distributions.
The above benchmark problems did not have such characteristics, and thus the effectiveness of the \textit{block} strategy could not be confirmed.

\subsubsection{Reinforcement learning}

To verify the effectiveness of the \textit{block} strategy, additional benchmarks for reinforcement learning (RL) problems, where the data generative distribution depends on an agent's policy.
Data with nonoptimal behaviors obtained by the past policy can be regarded as a kind of erroneous data, since they would cause a bias in agent's value functions.
If their replay can be appropriately blocked, the robust policy should be obtained, improving control performance.

Specifically, two tasks in OpenAI Gym, i.e. \textit{InvertedDoublePendulum-v4} (DoublePendulum) and \textit{Reacher-v4} (Reacher), are solved using a soft actor-critic (SAC) algorithm~\citep{haarnoja2018soft}.
The implementation of SAC was adapted from that used in the literature~\citep{kobayashi2025intentionally}.
The training timing is after four interactions, at which time up to one batch is replayed for training.
DoublePendulum and Reacher are trained with 1,500 and 1,000 episodes, respectively, followed by 100 episodes with the learned policies for evaluation.
In the RL problems, the interquartile mean (IQM) of the returns (a.k.a. the sum of rewards in episode) is computed as a score of each trial according to the literature~\citep{agarwal2021deep}.

The following four conditions are compared in the above tasks with 20 different random seeds.
\begin{itemize}
    \item \textit{FIFO}: only with FIFO buffer as like the standard RL algorithms with experience replay~\citep{lin1992self,isele2018selective}
    \item \textit{DER}: the conventional method
    \item \textit{-B}: without the \textit{block} strategy
    \item \textit{A2ER}: the proposed method
\end{itemize}
To ensure fairness, the buffer size in the case only with the FIFO buffer is set to $N^{\mathrm{FIFO}} \gets N^{\mathrm{FIFO}} + N^{\mathrm{RS}} = 1024$, and the batch size from it is also set to $B \times 2 = 64$.
Note that, even with the enlarged FIFO buffer, its size is still far smaller than those used in general RL implementations, and generalization performace is expected to be decreased since it cannot retain enough past information.
On the other hand, DER and the proposed method, which have RS buffers and regularization to past outputs, can improve generalization performance, but excessive consolidation of past outputs would inhibit learning because some past data become inappropriate for learning due to distribution shifts.

The learning results are summarized in Table~\ref{tab:result_a2er_rl}.
Here, as well as the above benchmarks, a weighted average based on the ranks of the scores is used to prioritize the worst cases and show robustness.
The first remarkable point is that DER made little progress in learning.
This is due to the fact that $\alpha=1$ was too large and the function to maintain past outputs was too strong, which prevented the value function from being updated, failing to optimize the policy.
In addition, FIFO, a common implementation of RL, had a poor task success rate with DoublePendulum, and in Reacher, its score did not reach a satisfactory level, while not failing significantly.
This result is as expected, and can be attributed to the generalization performance degradation caused by too small a buffer size.

On the other hand, the proposed method, A2ER, outperformed the other methods for both tasks.
Although Reacher's score does not seem to differ significantly from that of FIFO, both the worst and best cases of A2ER exceeded those of FIFO, indicating that the generalization performance of A2ER was stably obtained.
This result is largely due to the \textit{block} strategy, as a clear performance drop was observed in the case without it.
In particular, the number of failure cases increased for both tasks, with the worst Case more than twice as bad in Reacher.
Thus, without the \textit{block} strategy, it was confirmed that the influence of past erroneous data can cause instability in learning and a decrease in generalization performance.

From the above results, we can conclude that all three strategies in the proposed method, A2ER, are important to improve the performance of DER.

\begin{table}[tb]
    \caption{Results of RL problems for evaluating A2ER:
    the average of 20 trials in each condition was weighted by ranks, giving priority to the worst case.
    }
    \label{tab:result_a2er_rl}
    \centering
    \begin{tabular}{l|cc}
        \hline\hline
        Method & \multicolumn{2}{c}{RL (IQM of returns: larger is better)}
        \\
         & DoublePendulum & Reacher
        \\
        \hline
        FIFO & 873.66 & -12.29
        \\
         & [206.71, 9359.81] & [-13.05, -10.13]
        \\
        DER & 60.55 & -34.86
        \\
         & [35.53, 129.96] & [-52.19, -12.50]
        \\
        -B & 1905.71 & -16.08
        \\
         & [130.18, 9359.86] & [-27.75, -9.72]
        \\
        A2ER & \textbf{2896.43} & \textbf{-11.69}
        \\
         & [353.61, 9359.65] & [-12.51, -8.43]
        \\
        \hline\hline
    \end{tabular}
\end{table}

\subsection{Results of O2S}
\begin{figure}[tb]
    \centering
    \includegraphics[keepaspectratio=true,width=\linewidth]{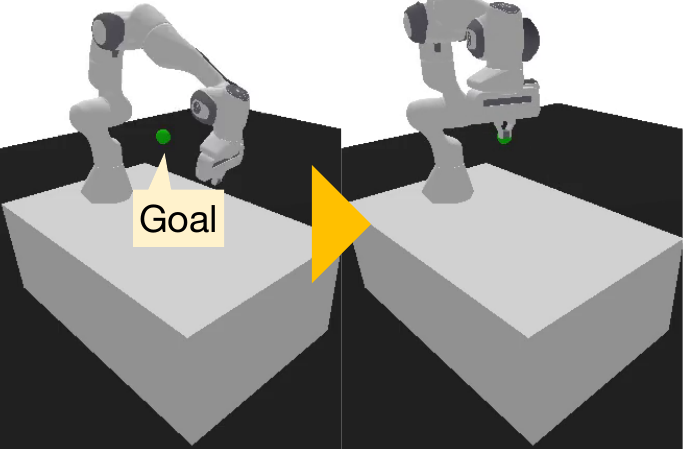}
    \caption{Snapshots of goal-conditioned RL}
    \label{fig:prob_gcrl}
\end{figure}

\begin{table*}[tb]
    \caption{Results of goal-conditioned RL problems for evaluating O2S:
    the average of 20 trials in each condition was weighted by ranks, giving priority to the worst case;
    top-2's methods were noted in bold.
    }
    \label{tab:result_o2s}
    \centering
    \tiny
    \begin{tabular}{l|ccc|ccc}
        \hline\hline
        Method & \multicolumn{3}{c|}{End-effector space} & \multicolumn{3}{c}{Joint space}
        \\
         & E+N-G & E-N+G & E+N+G & J+N-G & J-N+G & J+N+G
        \\
        \hline
        RS & -3.62 & -6.25 & -6.84 & -7.30 & -15.64 & -16.13
        \\
         & [-5.63, -1.70] & [-11.57, -3.21] & [-11.38, -3.52] & [-11.86, -4.36] & [-19.80, -8.02] & [-21.14, -11.71]
        \\
        Q2S & -2.47 & -3.97 & \textbf{-4.22} & -6.19 & -9.14 & \textbf{-8.13}
        \\
         & [-3.28, -1.17] & [-7.00, -2.40] & [-5.30, -3.22] & [-8.58, -3.83] & [-12.06, -5.19] & [-9.64, -5.81]
        \\
        P2S & \textbf{-2.25} & \textbf{-3.77} & -4.53 & \textbf{-5.03} & \textbf{-9.00} & -8.17
        \\
         & [-3.04, -1.34] & [-6.15, -1.94] & [-5.30, -2.85] & [-8.06, -3.70] & [-16.95, -4.19] & [-9.65, -5.55]
        \\
        O2S & \textbf{-2.36} & \textbf{-3.36} & \textbf{-4.11} & \textbf{-5.48} & \textbf{-7.64} & \textbf{-7.63}
        \\
         & [-2.81, -1.27] & [-4.63, -2.30] & [-5.48, -2.83] & [-8.53, -3.26] & [-9.31, -4.34] & [-9.20, -5.24]
        \\
        \hline\hline
    \end{tabular}
\end{table*}

Next, O2S, a modification of RS, is examined.
To this end, goal-conditioned RL~\citep{liu2022goal} problems are solved since they need both the memory consolidation and plasticity.
In other words, an agent needs to acquire the policy to achieve any goals while different goals are randomly specified in each episode.
If the implementation has only a (small) FIFO buffer and is too plastic, it will forget how to achieve the goals conditioned in the first half of the learning process.
Conversely, if the consolidation is too high, as in the case of ordinary RS, it will not be able to take into account the goals conditioned in the second half of the learning process.
While taking advantage of the ability to store old data, which is an advantage of the RS buffer, an appropriate balance should be achieved by moderately updating the buffer with new data and actively excluding unnecessary data.

The specific task is given as a modification of \textit{PandaReachDense-v3} in Panda-Gym~\citep{gallouedec2021panda} (see Fig.~\ref{fig:prob_gcrl}).
There are three major modifications.
First, the robot's initial joint angles can now be randomized using a uniform distribution.
Second, the goal position to be reached by the robot's end-effector is offset vertically by the length of the fingertips, and the range of the uniform distribution can be specified.
Third, when the robot's joint space is specified as the action space, a penalty for the orientation error of the fingertip is added to the reward.
In addition, the end-effector's orientation and joint angles are added to the state space (originally, the end-effector's position and velocity) to accommodate this extension.
Under these modifications, the task is trained over 10,000 episodes in SAC with A2ER as in the RL problems above.
The task is divided into three types by adding a uniform noise with a width of $\pi/4$ to the initial joint angles (N) and/or a uniform noise with a width of $0.4$ to the goal position (G).
Furthermore, the action space can be divided into two types depending on whether it is a 3-DOF end-effector position space (E) or a 7-DOF joint space (J), resulting in a total of six types.

Since the three strategies in O2S are implemented in stages, it is not possible to perform ablation tests as was doone in the benchmarks above.
Instead, they are added to the original RS step by step to confirm the usefulness of each strategy.
That is, the following four conditions are tested.
\begin{itemize}
    \item \textit{RS}: the conventional method
    \item \textit{Q2S}: only with the \textit{q-logarithm} strategy
    \item \textit{P2S}: without the \textit{omission} strategy
    \item \textit{O2S}: the proposed method
\end{itemize}
Note that P2S and O2S introduce the \textit{plural} strategy with $L=2$, so the two RS buffers are with $N^{\mathrm{RS}} = 256$ each.

The results of 20 trials of each of the above conditions with different random seeds are summarized in Table~\ref{tab:result_o2s}.
The evaluation way is the same as for the RL problems above.
The basic trend was that the joint space is more difficult than the end-effector position space, resulting in lower returns (partly due to the inclusion of the orientation error penalty).
With the noise added only to the initial joint angles, less than 10,000 episodes would be sufficient for learning, since similar situations might be visited in the trial-and-error process.
On the other hand, the difficulty was increased with the noise added to the goal position, and in that time, the conventional RS could not acquire the optimal policy to reach multiple goals.

In contrast, adding the \textit{q-logarithm} strategy to suppress the decay of the acceptance probability of new data into RS buffer allowed more goals to be stored in the buffer(s), resulting in a clear performance improvement.
The performance improvement with the \textit{plural} strategy appears to be minor.
However, focusing on the best results, there was a clear improvement in the cases with the randomized goals.
That is, although the \textit{plural} strategy did not stabilize the learning enough, it improved the expected scores.

Finally, when the \textit{omission} strategy was added, the top-2 was achieved in all task conditions.
In particular, there was a significant improvement in performance when the goal was randomized.
In addition, the worst-case results in almost all conditions were improved, indicating that learning was successfully stabilized.
The reason of these improvements seem not to be the increase of plasticity as like the \textit{q-logarithm} and \textit{plural} strategies.
In fact, we can find the final acceptance probability of $L$-th RS buffer, as summarized in Table~\ref{tab:result_prob}.
From this, RS clearly lacked plasticity, and it acquired plasticity with the \textit{q-logarithm} strategy.
The \textit{plural} strategy also increased plasticity (given that the first RS buffer accepted more data).
On the other hand, the increase in the acceptance probability due to the addition of the \textit{omission} strategy was small, suggesting no change of plasticity.
Therefore, it is implied that the \textit{omission} strategy contributed to the selection of data, and actively removing unnecessary data from the buffer promoted more stable learning.

From the above results, we can conclude that all three strategies in the proposed method, O2S, are important to improve the performance of RS.

\begin{table}[tb]
    \caption{Final acceptance probability of $L$-th RS buffer}
    \label{tab:result_prob}
    \centering
    \begin{tabular}{ll|c}
        \hline\hline
        Method & Condition & Probability
        \\
        \hline
        RS & $L=1$, $q=0$ & 0.10
        \\
        Q2S & $L=1$, $q=1$ & 12.69
        \\
        P2S & $L=2$, $q=1$, $\zeta=0$ & 13.32
        \\
        O2S & $L=2$, $q=1$, $\zeta=0.2$ & \textbf{13.43}
        \\
        \hline\hline
    \end{tabular}
\end{table}

\section{Conclusion}

This paper proposed A2ER and O2S, which introduce three improvement strategies for DER and RS, respectively, a versatile baseline for continual learning that does not require task label information.
The conventional methods mainly aim to maintain past data and outputs, i.e. memory consolidation, which in turn impairs memory plasticity, i.e. quick adaptation to new data, since they are in a tradeoff relationship.
Therefore, the proposed methods made the improvements in order to achieve a better balance between consolidation and plasticity.
The numerical results indicated that all strategies steadily improve learning performance in benchmarks where not only consolidation but also plasticity are important.

As future work, as the proposed method introduces several hyperparameters, automatic adjustment of them is required.
In this regard, it might be important to determine whether the task at hand requires consolidation or plasticity, and to determine with higher accuracy which unnecessary data should not be replayed or stored.
By acquring them, the proposed method will be able to be applied to the training of larger-scale practical models with numerous data.
In particular, the proposed method is expected to be useful in areas where data is likely to increase continuously, such as recent robotic foundation models~\citep{firoozi2023foundation} and analysis of various human motions~\citep{kong2022human}.

\backmatter

%
%
%

\bmhead{Acknowledgments}

This research was supported by ``Strategic Research Projects'' grant from ROIS (Research Organization of Information and Systems).

\section*{Declarations}

\bmhead{Competing Interests}

The author declares that there is no known competing financial interests or personal relationships that could have appeared to influence the work reported in this paper.

\bmhead{Author Contributions}

Taisuke Kobayashi contributed to everything for this paper: Conceptualization, Methodology, Software, Validation, Investigation, Visualization, Funding acquisition, and Writing.

\bmhead{Compliance with Ethical Standards}

The data used in this study was exclusively generated by the author.
No research involving human participants or animals has been performed.

\bmhead{Data Availability}

The datasets generated during and/or analyzed during the current study are available from the corresponding author on reasonable request.

\begin{appendices}

\section{Influence of quantile operation}
\label{app:quantile}

\begin{table*}[tb]
    \caption{Results of toy problems for evaluating the effects of $\rho$ in the quantile operation:
    the average of 20 trials in each condition was weighted by ranks, giving priority to the worst case;
    top-2's methods were noted in bold.
    }
    \label{tab:result_quantile}
    \centering
    \tiny
    \begin{tabular}{l|cccc|cccc}
        \hline\hline
        Method & \multicolumn{4}{c|}{Regression (KLD: smaller is better)} & \multicolumn{4}{c}{Classification (ACC: larger is better)}
        \\
         & R1 & R2 & R3 & R4 & C1 & C2 & C3 & C4
        \\
        \hline
        DER & 0.84 & 4.00 & 7.24 & 2.23 & 84.12 & 73.25 & 80.97 & 84.58
        \\
         & [0.64, 0.92] & [3.71, 4.18] & [6.81, 7.48] & [1.98, 2.32] & [82.69, 85.82] & [71.37, 75.64] & [77.81, 83.91] & [82.70, 86.82]
        \\
        $\rho=0.25$ & 0.56 & 3.33 & 9.14 & 1.83 & 84.08 & 70.46 & 82.82 & 89.66
        \\
         & [0.33, 0.84] & [2.41, 4.81] & [4.52, 12.12] & [1.54, 2.12] & [83.16, 86.71] & [69.00, 75.01] & [81.48, 87.23] & [88.35, 92.40]
        \\
        $\rho=0.5$ & \textbf{0.42} & \textbf{2.87} & \textbf{7.03} & \textbf{1.71} & \textbf{91.26} & \textbf{87.52} & \textbf{91.28} & \textbf{95.82}
        \\
         & [0.30, 0.49] & [2.37, 3.59] & [4.44, 10.68] & [1.39, 1.78] & [90.07, 92.70] & [87.03, 88.70] & [89.89, 93.39] & [95.33, 96.85]
        \\
        $\rho=0.75$ & \textbf{0.36} & \textbf{2.93} & \textbf{6.49} & \textbf{1.68} & \textbf{91.29} & \textbf{87.66} & \textbf{91.14} & \textbf{95.49}
        \\
         & [0.24, 0.49] & [2.29, 3.33] & [4.20, 9.89] & [1.45, 1.78] & [90.67, 92.43] & [87.09, 89.35] & [89.82, 93.24] & [94.52, 96.68]
        \\
        \hline\hline
    \end{tabular}
\end{table*}

The dynamic threshold for the error from previous outputs required for the \textit{adaptation} strategy in A2ER strongly depends on the quantile $\rho \in [0, 1]$.
In addition, $\rho$ is also involved in calculating the correction rate in the \textit{correction} strategy, which affects the \textit{block} strategy as well.
Therefore, $\rho$ might be a very important hyperparameter that is involved in all A2ER strategies.

The toy problems performed in Section~\ref{subsec:res_a2er} are tested with three $\rho = 0.25, 0.5, 0.75$ (a.k.a. quartile points $Q_1/4$, $Q_2/4$, and $Q_3/4$, respectively).
The results are summarized in Table~\ref{tab:result_quantile}.
As can be seen in this table, $\rho=0.25$ was too small and the improvement from the conventional DER was not significant, but $\rho=0.5, 0.75$ improved clearly.
In addition, the performance difference between $\rho=0.5$ and $\rho=0.75$ was minor.
Hence, we can conclude that A2ER improves learning performance as expected when $\rho$ is set to some large value.
Based on this fact, $\rho=0.5$, which corresponds to the median value, is adopted in this paper for simplicity.

\section{Other possible design of counter}
\label{app:design}

The new generalized counter given by eq.~\eqref{eq:qlog} is soley a heuristic design, and other designs are considerable.
In that case, $0 \leq \Delta f(n) \leq 1$ must be satisfied, but one other important factor is whether it saturates with $\Delta f(n\to\infty) \to 0$.
That is, if the counter saturates, the probability that the RS buffer accepts new data always remains, and plasticity is ensured.
Conversely, if the counter does not saturate, the RS buffer will eventually accept almost no new data, thus ensuring consolidation.

With this in mind, two simple candidates are proposed.
First design changes only the slope when $n>N^{\mathrm{RS}}$.
\begin{align}
    f_q^\mathrm{lin}(n) &= \min(n, N^{\mathrm{RS}})
    \nonumber \\
    &\quad + \left\lfloor (1 - q) \max(0, n - N^{\mathrm{RS}}) \right\rfloor
\end{align}
where $q \in [0, 1)$.
In this design, the counter diverges to infinity without saturation for $q<1$.

The second is basically a saturating design, which corresponds to exponential decay RS~\citep{cormode2009forward,osborne2014exponential}.
\begin{align}
    f_q^\mathrm{exp}(n) &= \min(n, N^{\mathrm{RS}})
    \nonumber \\
    &\quad + \left\lfloor \frac{N^{\mathrm{RS}}}{q} \left\{1 - \exp\left( -\frac{q \max(0, n - N^{\mathrm{RS}})}{N^{\mathrm{RS}}} \right) \right\} \right\rfloor
\end{align}
where $q \in (0, 1])$.
Note that $f_{q\to0}^\mathrm{exp}(n)$ converges to $n$.
This design saturates at $q>0$ and the acceptance probability converges to $0.5$ when $q=1$ as in eq.~\eqref{eq:qlog} with $q=2$.

These designs including the \textit{q-logarithm} strategy are compared on the classification problem addressed in Section~\ref{subsec:res_a2er}.
However, the target function (i.e. the arrangement of the Gaussian mixture distribution) is switched between the first five cycles and the second five cycles in order to check the balance between consolidation and plasticity.
That is, excessive consolidation inhibits learning in the second half, while excessive plasticity fails even the first half.
Therefore, the classification accuracies of the first and second halves suggest consolidation and plasticity, respectively.
In addition, conditions and trials in which both accuracies exceed a certain level (in this case, 90~\%) are judged as an appropriate balance.

\begin{figure}[tb]
    \begin{subfigure}[b]{0.96\linewidth}
        \centering
        \includegraphics[keepaspectratio=true,width=\linewidth]{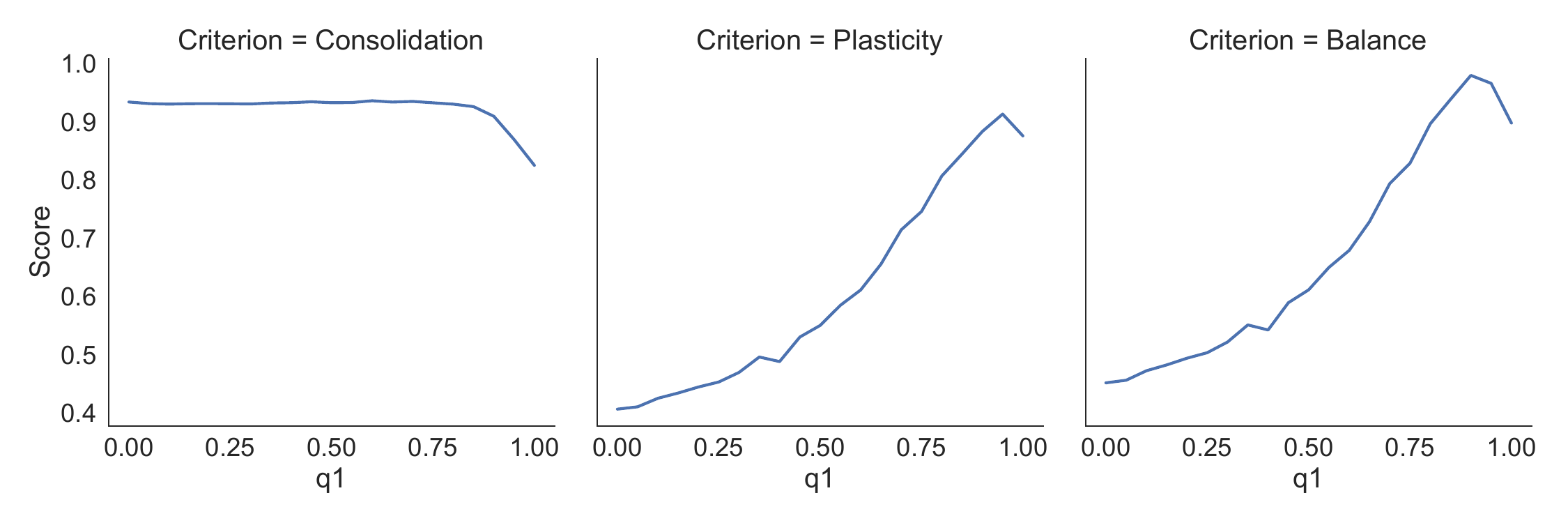}
        \subcaption{$f_q^\mathrm{lin}(n)$}
        \label{fig:summary_lin}
    \end{subfigure}
    \begin{subfigure}[b]{0.96\linewidth}
        \centering
        \includegraphics[keepaspectratio=true,width=\linewidth]{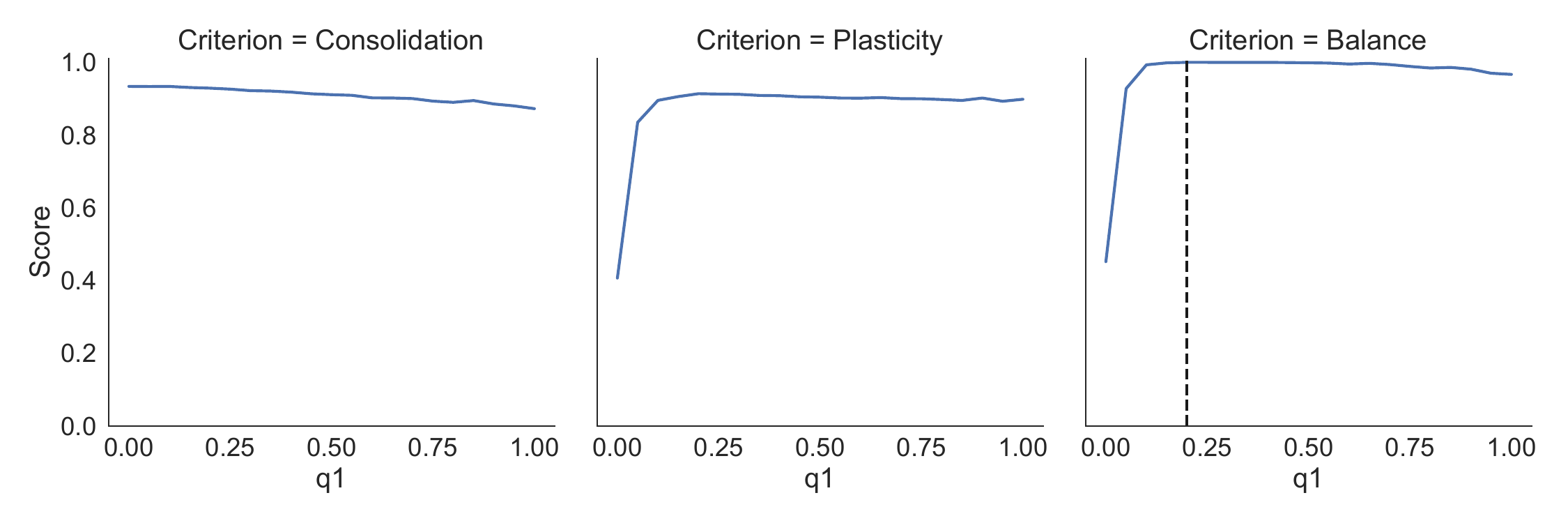}
        \subcaption{$f_q^\mathrm{exp}(n)$}
        \label{fig:summary_exp}
    \end{subfigure}
    \begin{subfigure}[b]{0.96\linewidth}
        \centering
        \includegraphics[keepaspectratio=true,width=\linewidth]{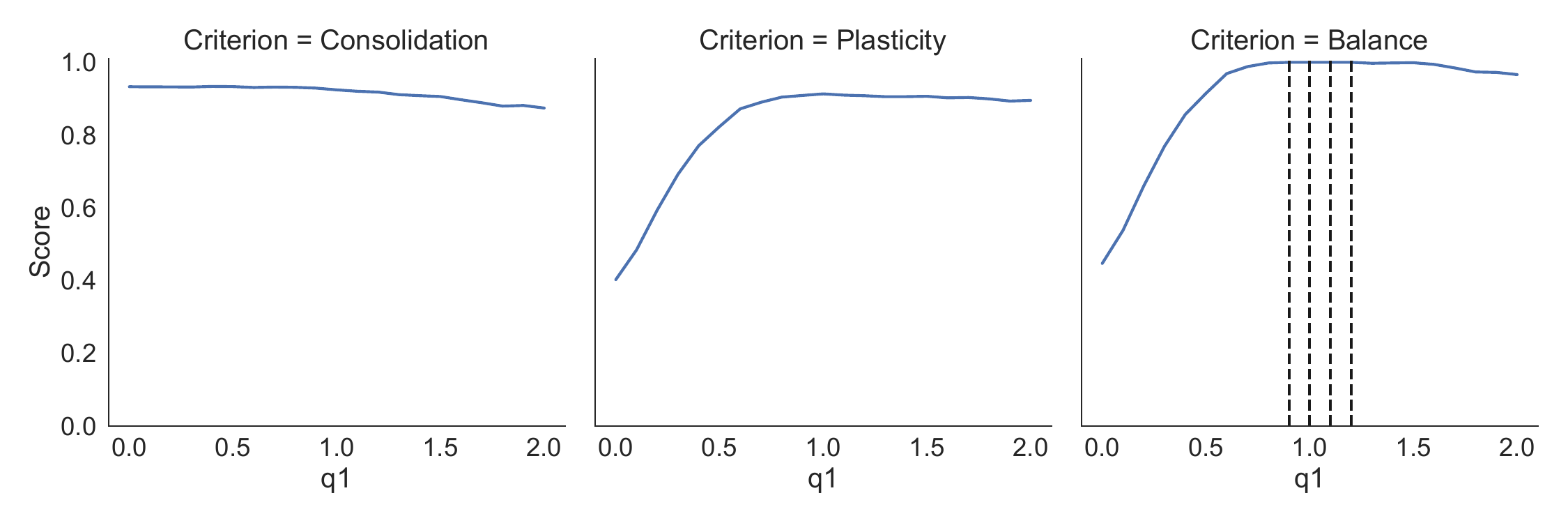}
        \subcaption{the \textit{q-logarithm} strategy $f_q(n)$}
        \label{fig:summary_lnq}
    \end{subfigure}
    \caption{Comparison of three possible designs for the generalized counter through the switched classification problem}
    \label{fig:summary}
\end{figure}

The hyperparameter $q$ for each design is divided into 21 equal parts, and the results of 20 trials for each condition are summarized in Fig.~\ref{fig:summary}.
As in the above experiment, the statistics here are rank-weighted averages, giving priority to the worst results.
The vertical dashed lines in the graphs on the right refer to the conditions in which the appropriate balance were achieved in all 20 trials.
In the case of $f_q^\mathrm{lin}(n)$ without saturation, the appropriate balance was never obtained due to the lack of plasticity.
The case of $f_q^\mathrm{exp}(n)$ with saturation had high levels of both consolidation and plasticity, but only one optimal hyperparameter was found.
The reason for the high level of consolidation, which was expected to be low, may be that the classification problem conducted did not require much consolidation.

On the other hand, the proposed \textit{q-logarithm} strategy found four optimal hyperparameters with $q=0.9, 1, 1.1, 1.2$.
In particular, $q=0.9, 1$ are not saturated, while $q=1.1, 1.2$ are saturated, suggesting that the appropriate balance is hidden around the boundary.
Since the proposed method can specify the presence or absence of saturation by hyperparameters, the appropriate balance between consolidation and plasticity can be robustly obtained with a small dependence on hyperparameters.
Therefore, this design was adopted in this paper, and the hyperparameter is set to $q=1$, which is the boundary without saturation.
Note that the \textit{plural} strategy enables to have saturation in the shallower buffers with $q > 1$ (in this paper, the first buffer with $q=1.5$).

\end{appendices}

\bibliography{biblio}

\end{document}